# Explicit Computation of Input Weights in Extreme Learning Machines

Jonathan Tapson[1], Philip de Chazal[1] and André van Schaik[1]

[1] The MARCS Institute, University of Western Sydney, Penrith NSW, Australia 2751

{j.tapson, p.dechazal, a.vanschaik}@uws.edu.au

**Abstract.** We present a closed form expression for initializing the input weights in a multilayer perceptron, which can be used as the first step in synthesis of an Extreme Learning Machine. The expression is based on the standard function for a separating hyperplane as computed in multilayer perceptrons and linear Support Vector Machines; that is, as a linear combination of input data samples. In the absence of supervised training for the input weights, random linear combinations of training data samples are used to project the input data to a higher dimensional hidden layer. The hidden layer weights are solved in the standard ELM fashion by computing the pseudoinverse of the hidden layer outputs and multiplying by the desired output values. All weights for this method can be computed in a single pass, and the resulting networks are more accurate and more consistent on some standard problems than regular ELM networks of the same size.

# 1 Introduction

The Extreme Learning Machine has proven its usefulness as a fast and accurate method for building classification and function approximation networks [1]. Its usefulness stems in large part from the fact that it requires no incremental or iterative learning, and has no free parameters which need to be tuned to get the optimal results. Naïve application of a standard ELM to most benchmark problems produces results which are significantly better than most results from highly-tuned iterative methods. In all but a few applications, the ELM can only be outperformed by time-consuming and expertly-applied techniques such as multiple layer deep-learning networks, and recent extension of the ELM to multiple layers suggests that they have the potential to outperform deep learning networks too [2].

One area which has been identified as offering some potential for improvement in ELM is the specification of the input weights, which connect the input neurons to the hidden layer neurons. In a standard ELM these are generally initialized to random values from a uniform distribution on some appropriate range, and thereafter they remain fixed throughout the use of the ELM.

The random input layer projection implemented in ELM is a contrast to almost all other machine learning techniques, which use supervised learning to arrive at explicit values for the projections at all layers (or in the case of support vector machines (SVM), to choose the support vectors [3]). In this report we proceed from the knowledge that for both multi-layer perceptrons (MLP) trained by standard backpropagation techniques (backprop), and for SVM, the weights produce a projection which is a linear combination of training data samples [3, 4]. We suggest a technique in which randomized linear combinations of input data samples can be systematically produced to provide the input layer weights for ELM, and demonstrate that in some traditionally difficult classification problems, this method results in a superior performance to standard ELM.

# 2 Weights and Training Samples

The standard ELM is often described as an MLP with an input layer, a hidden layer of neurons with nonlinear activation functions, and an output layer with linear activation functions. The outputs can be described by:

$$y_{n,t} = \sum_{j=1}^{d} {w_{nj}}^{(2)} g\left(\sum_{i=1}^{k} {w_{ji}}^{(1)} x_{i,t}\right) \quad (1)$$

where $y_t$ is the output layer vector corresponding to input data $x_t$, and $t$ is the sample index ($t$ would be the time step index for time series data). The output $y_t$ is a line-ar sum of hidden layer outputs weighted by $w_{nj}$. There are $k$ input neurons and $d$ hidden layer neurons; $i$ is the input layer index, $j$ the hidden layer index and $n$ the output layer index. The hidden layer neurons have a nonlinear activation function $g(\cdot)$ which acts on the sums of linearly weighted inputs $w_{ji}x_i$. The superscripts indicate the layer. If the outputs $y_{n,t}$ were also to be acted on by a nonlinear activation function like $g(\cdot)$, this would be a conventional three-layer MLP. Note that in the development here we are assuming that any bias term would be present in the training data as an input data element.

In a standard ELM the input layer weights are initialized to random values, uniformly distributed in some sensible range – say (*-1, 1*). In this report we describe a closed-form solution for determining these weights.

Consider the conventional backpropagation of error algorithm in MLPs [4]. The training algorithm contains two phases. In the forward phase, the error for each training sample is calculated. In the backward phase, the weights for each layer are updated by multiplying the error for that layer by the activations of that layer (which produces an error gradient) and subtracting a fraction of the gradient from the weights. So, in the network given above, the update for the input layer weights would be of the form

$$\Delta {w_{j,t}}^{(1)} = \alpha \cdot {E_t}^{(1)} \cdot x_t \tag{2}$$

where $\alpha$ is the learning rate (which defines the fraction of the gradient used in updates), and $E$ is the error gradient. It can be seen that each update consists of adding or subtracting some fraction of an input training sample to the weights. If the weights are initialized to zero (which is not unreasonable) then the final value of the weights will be some linear combination of the input samples; for $h$ training samples, after one pass through all training data,

$${w_j}^{(1)} = \alpha \sum_{t=1}^{h} {E_t}^{(1)} \cdot x_t \tag{3}$$

We find a similar result for the support vector machine [3]. In the linear case, an SVM classifier is designed to find the maximum-margin separating hyperplane between two classes of training data. This optimization problem is generally expressed as follows: given that the separating hyperplane can be expressed as an inner product in the input space such that

$$w \cdot x - b = 0 \tag{4}$$

where $b$ is a bias term, so that the hyperplane does not have to pass through the origin (note that here $w$ is a generic weight vector, not related to $w$ in (1) - (3); we persist with the notation in order to emphasize the commonality in these expressions). The classification margins to this separating hyperplane can be expressed as two parallel hyperplanes at

$$w \cdot x - b = \pm 1 \quad (5)$$

We need to maximize the distance $2/\|w\|$ between these two planes, subject to the constraint that

$$y_t(w \cdot x_t - b) \geq 1 \quad (6)$$

for all *t*. This constraint requires that all samples remain outside the margins, as is implicit in their meaning. According to the Karush-Kuhn-Tucker condition [5], the solutions can be expressed as a linear combination of input samples

$$w = \sum_{t=1}^{h} \alpha_t \cdot y_t \cdot x_t \quad (7)$$

Note that $\alpha$ here is not the same as the learning rate in the MLP expression above, but is a multiplication factor (a Lagrange multiplier in the optimization process). We have used the same symbol $\alpha$ as this symbol is conventional in both usages in MLP and SVM and serves to emphasize the point of this development, which is that in both MLP and linear SVM cases the weight vectors are linear weighted sums of the input training samples (note the similarities between (3) and (7) above). This should come as little surprise given the acknowledged equivalences between perceptrons and linear SVMs, and MLPs and SVMs (see for example Collobert and Bengio [6]). Similarly, it is consistent with the intuition that the dot product between vectors (such as a training sample and a testing sample) is a measure of their similarity.

Given that the optimal input layer weights in MLPs and the support vectors in SVMs are linear functions of the training samples, it might be the case that an ELM in which the input layer weights were biased towards linear combinations of the input training samples would perform better than one in which the input layer weights were uniformly distributed random weights. There are a number of questions to be addressed here:

- Does an ELM in which the input layer weights are biased towards the input training samples perform better than one with uniformly distributed random values?
- How would we achieve the biasing of the input layer without any of the tedious incremental or stochastic learning of weights which ELM so successfully avoids?
- Is there a rigorous theoretical basis for the belief that an ELM would perform better with appropriately biased weights?

In the work reported here we propose some answers to the first two points, by suggesting a fast closed-form expression for generation of weights biased towards the input training samples, and showing that in some important cases it consistently outperforms a conventional ELM.

## 3 Methodology

In the method described here, which we refer to as Computed Input Weights ELM (CIW-ELM), we want to generate input weights which have the form of (3) or (7) above; that is to say, they must be (random) weighted sums of the training data samples. The input data are generally normalized to have zero mean and unity standard deviation, and weights are normalized to unity magnitude. ELM hidden layers are often specified in terms of number of neurons, or as a multiplier of the size of the input layer. We divide the number of hidden layer neurons *d* by the number of classes *C*, and generate weights for each group of $p = d/C$ hidden layer neurons from input vectors of a single class (we can reasonably define *d* so that $d/C$ gives an integer number of neurons). We then produce random weights biased to the training samples for the class, by summing the inner products of random binary-valued vectors with the training samples, according to the formula

$$w^{(1)} = R \cdot x \qquad (8)$$

where $x_{h \times k}$ is the full set of training data for the class, and $R_{p \times h}$ is a matrix of random binary values (random sign values), i.e. with elements having values in {-1, 1}. Should repeatable weight generation be required, the elements of *R* could be generated from well-known pseudorandom sequences such as Gold codes, thereby enabling a deterministic weight generation schema. Use of orthogonal pseudorandom codes would also guarantee orthogonality of the input weights, which has been identified to improve the generalization of ELM networks [2, 7].

The outcome of this method is that the weight $w_{ji}^{(1)}$ connecting input *i* with hidden layer neuron *j* is the sum of the input sample elements $x_i$ (where *i* is the vector index, i.e. $x_i$ is the $i^{th}$ element of any training sample *x*), for all the training samples of one class, where each individual element has had its sign flipped or preserved with random probability.

The columns of weight matrix $w^{(1)}$ are then normalized to unit vectors:

$$w_a{}^{(1)} = \frac{R_a}{|R_a|} \qquad (9)$$

where *a* is the column index.

In practical terms, the weights are computed according to the following algorithm:

1. Normalize all training data, so as to avoid use of scaling factors.

2. Divide the $d$ hidden layer neurons into $C$ blocks, one for each of $C$ output classes; for data sets where the number of training data samples for each class are equal, the block size is $B = d/C$. We denote the number of training samples per class as $K$. Where the training data sets for each class are not of equal size, the block size can be adjusted to be proportional to data set size.
3. For each block, generate a random matrix $R_{B\times K}$ of signs.
4. Multiply the input training data set for that class, $x_{K\times k}$, by $R$ to produce $B\times k$ summed inner products, which are the weights for that block of neurons.
5. Normalize the weights to unity magnitude as shown in (9).
6. Concatenate these $C$ blocks of weights for each class, each $B\times k$ in size, into one weight matrix $w_{d\times k}$.
7. Solve for the hidden layer weights of the ELM using standard ELM methods such as singular value decomposition on the computed hidden layer outputs.

## 4 Results

We have used the CIW-ELM method on several benchmark problems in machine learning. Its performance is perhaps best illustrated in terms of the well-known MNIST handwritten digit recognition (OCR) problem [8]. MNIST is a particularly challenging data set for classification as the dimensionality of the data is high (784 pixels for input, 10 classes for output, and 60000 training samples). We have previously reported good results with conventional ELM on MNIST [9]. Here we show that the use of CIW-ELM gives significantly greater accuracy for similar-sized networks than conventional ELM (see Fig. 1).

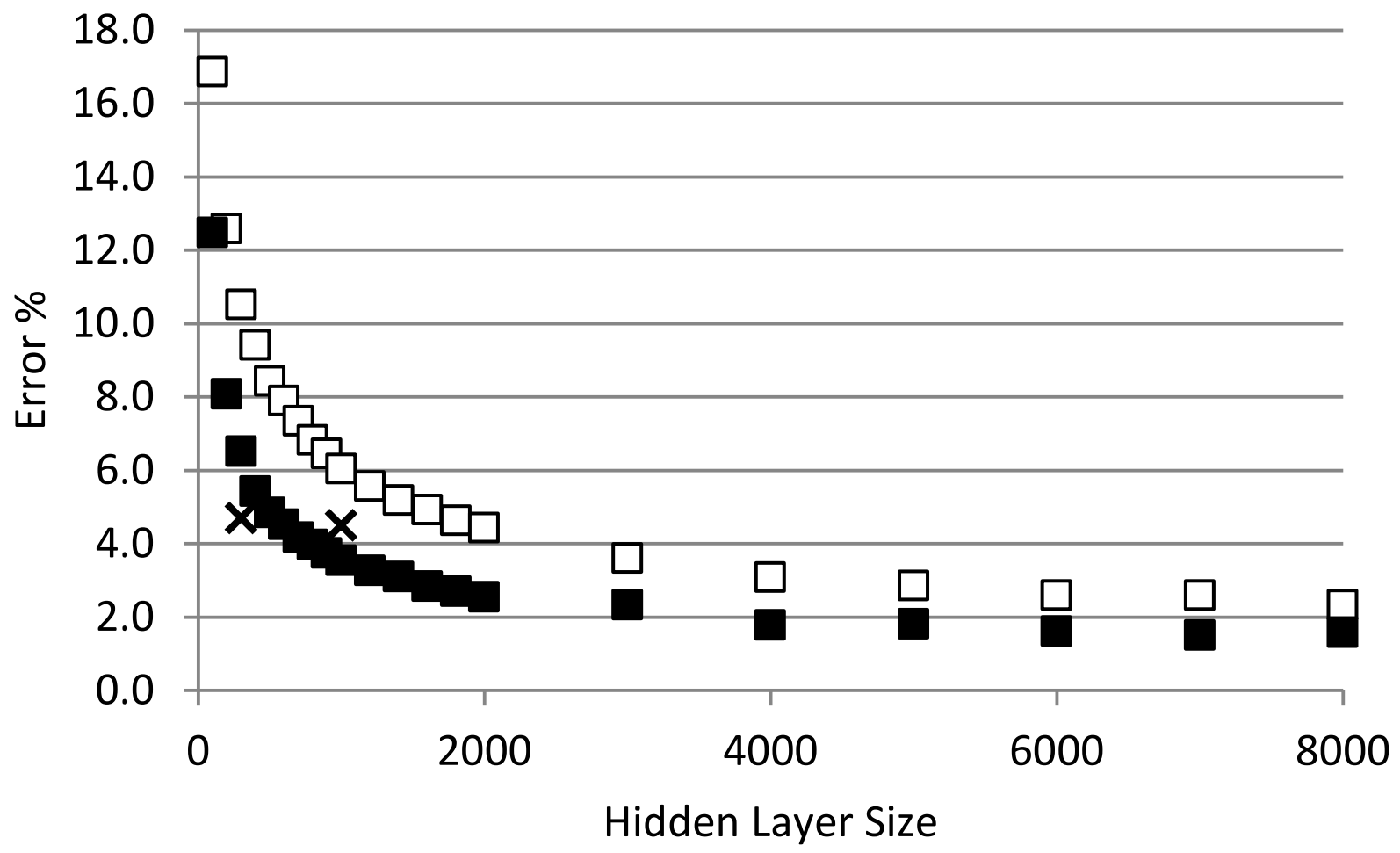

**Fig. 1.** Accuracy of standard ELM (white squares) and CIW-ELM (black squares) on the MNIST data set, for various hidden layer sizes. It can be seen that CIW-ELM consistently produces more accurate results for a given hidden layer size. Input data were not deskewed or centered, or in any way preprocessed except for normalization. The pseudoinverse solution was not regularized. The points indicated by the two crosses are MLP results from LeCun et al. [8] for 784-300-10 and 784-1000-10 networks, which represent the best results likely for back-propagation on equivalent networks (backpropagation is of course much more time-consuming than either ELM technique here).

One of the most significant advantages of the ELM method is in speed of implementation, because there is no incremental learning or optimization required. The most time-consuming computation in ELM is the calculation of the Moore-Penrose pseudoinverse solution to output weights, and this computation generally scales as $d^2$ where $d$ is the hidden layer size. For very large datasets, such as MNIST and in modern "big data" problems, the size of the pseudoinverse computation can become problematic. As such, the potential to use CIW-ELM to obtain high accuracy with smaller hidden layers has real benefits. This is illustrated in Figure 2 which compares execution times for the ELM and CIW-ELM; note that these were carried out on a very modest computational platform, illustrating the high speed of ELM implementation.

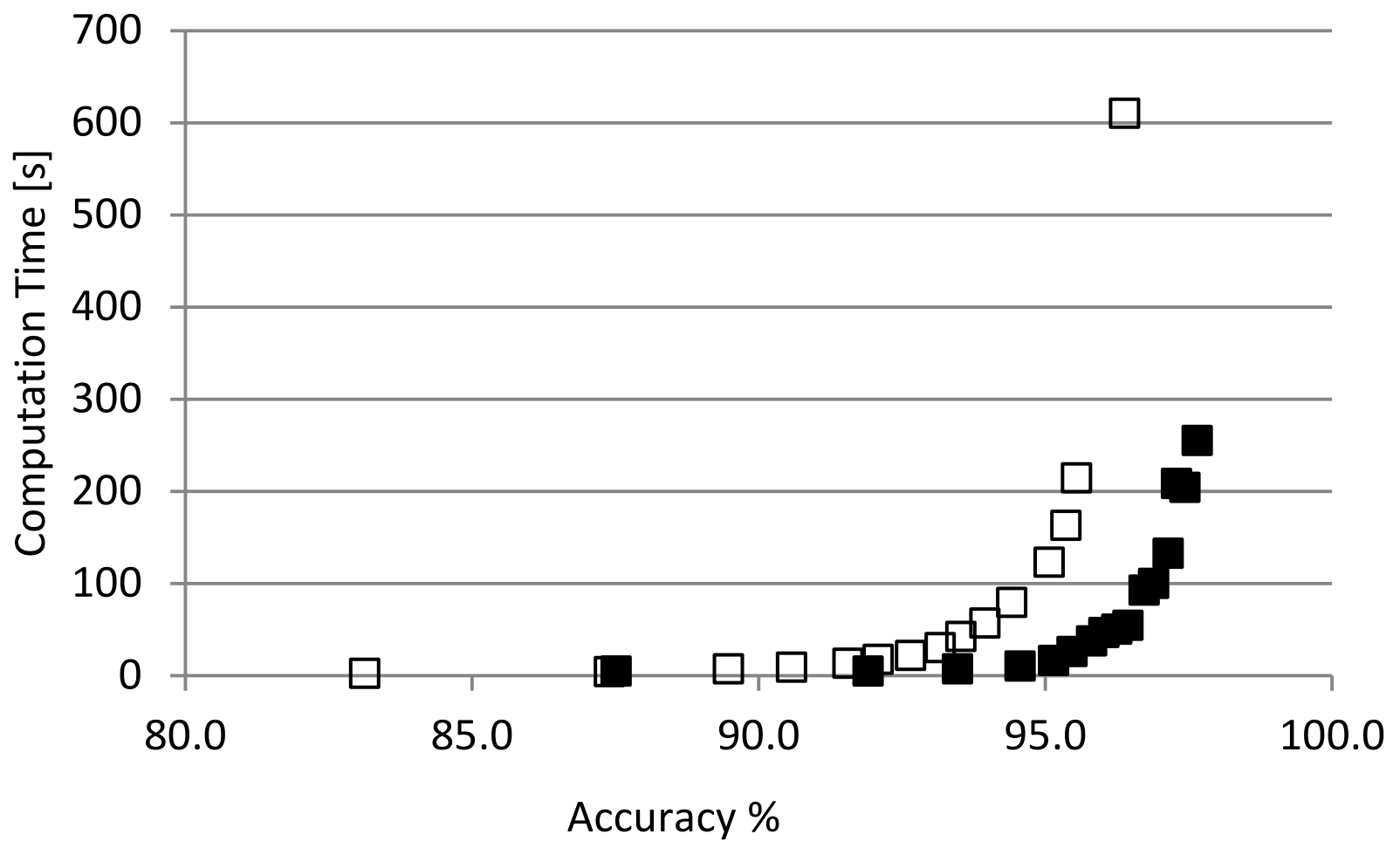

**Fig. 2.** Computation time required to reach a particular degree of accuracy on the MNIST OCR problem, for regular ELM (white squares) and CIW-ELM (black squares). The computation involves implementation (training) and testing of the network, including computation of input layer weights in CIW-ELM, and pseudoinverse solution of output weights for all 60 000 training cases. The reduction in time for CIW-ELM is a function of the smaller network size for a given accuracy; for example, ELM requires a 784-3000-10 network to achieve 96% accuracy, whereas CIW-ELM achieves this with a 784-700-10 network. Times were measured for a MATLAB R2012B implementation running on a 2010 MacBook Air 4 with an Intel dual core i5 processor clocked at 1.6GHz.

We have also used CIW-ELM on other standard classification problems, such as the well-known *abalone*, *iris*, and *wine* data sets [10]. The performance of CIW-ELM was superior to ELM for small network sizes in *wine* and *iris*, with convergence between the two methods at larger network sizes. There was no clear difference between the two methods for the *abalone* classification problem, but both methods produced results similar to the state of the art. Results for these data sets are illustrated in Figures 3-5.

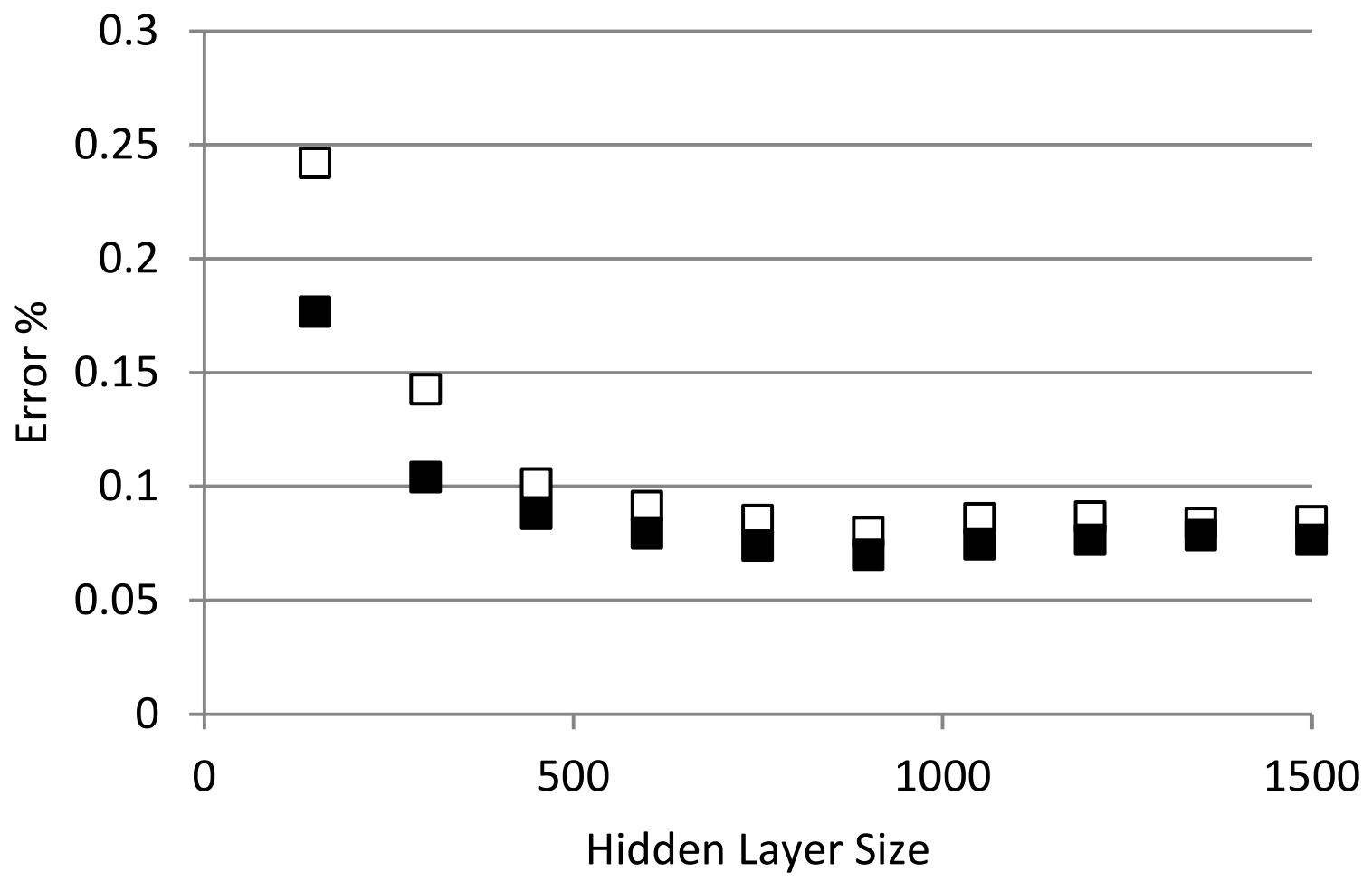

**Fig. 3.** Comparison of ELM (white squares) and CIW-ELM (black squares) on the *iris* classification problem. It can be seen that CIW-ELM shows more accuracy at all network sizes. There is some suggestion of overfitting for network sizes above 5-900-3, which is not surprising given the low dimensionality of the data (150 × 5 training data elements).

## 5 Discussion

The results illustrated in Figs. 1-5 suggest that for certain problems, the method may offer considerable improvement in performance over standard ELM. In cases where the data dimensionality is small, the performance of standard ELM converges rapidly onto the CIW-ELM result (however, it is for large data sets, where computation time becomes an issue, that the method shows the most promise). It is not yet clear whether the CIW-ELM method will improve results in function approximation tasks, as the algorithm as currently specified is intrinsically based on class information; we could in principle treat all input data as a single case for the purpose of a regression implementation. It is significant that the only problem so far on which CIW-ELM has not convincingly outperformed standard ELM is the *abalone* problem, which is really a regression problem (estimate the age of the abalone) recast as a classification problem (bin the abalone into one of three age bins). We note that Huang has demonstrated that SVM intrinsically requires the bias term $b$ in (4) – (6) and that this forces optimization within a more constrained space than standard ELM, which was designed from the beginning for regression or function optimization [11]. Our use of expressions (3) and (7) as models for the input weights will similarly have constrained the solution space, and this may also contribute to a lack of improvement in regression problems, although this may be difficult to evaluate.

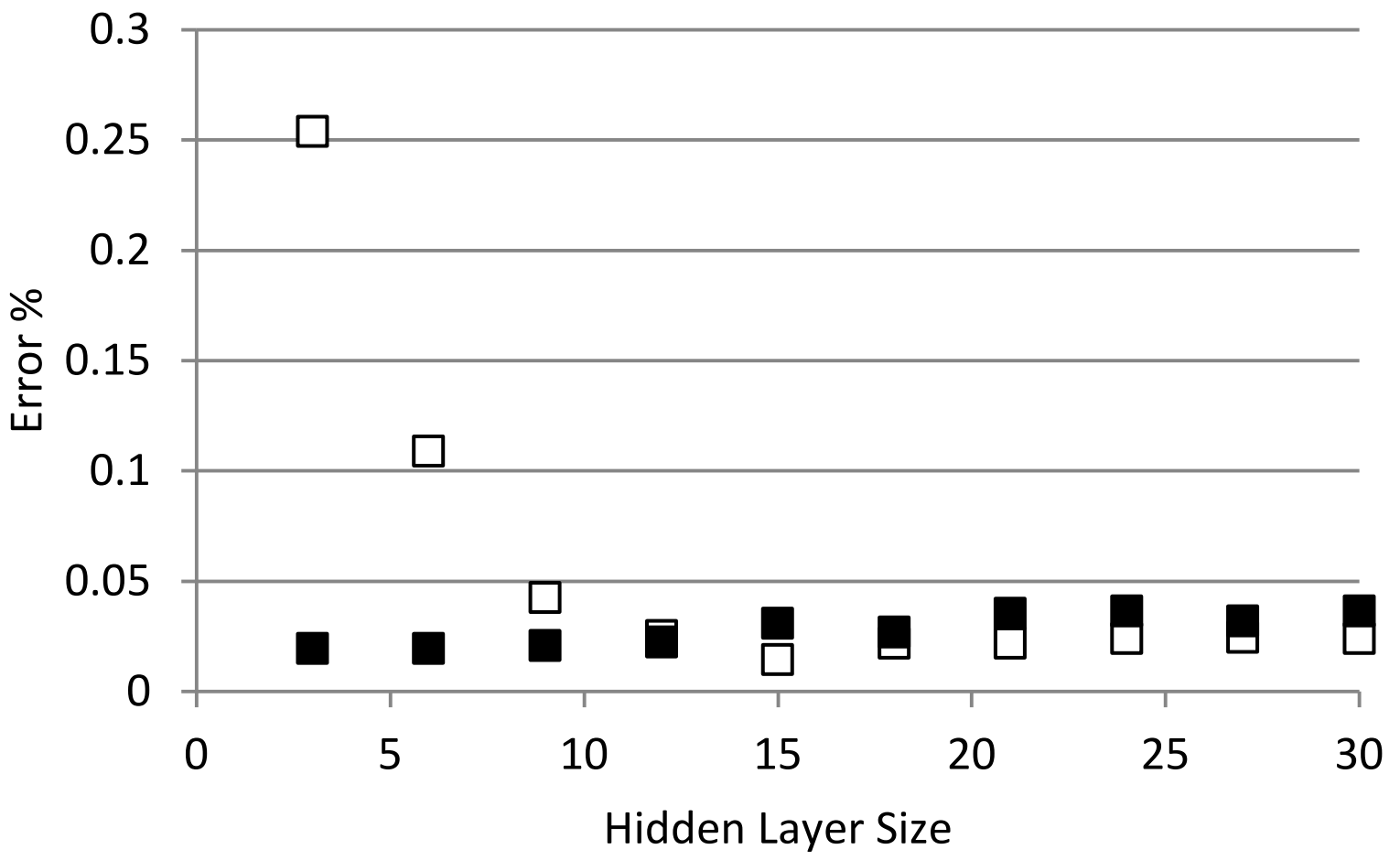

**Fig. 4.** Comparison of ELM (white squares) and CIW-ELM (black squares) performance on the *wine* classification benchmark. It can be seen that CIW-ELM performs extremely well for small networks, with both methods showing overfitting for larger network sizes.

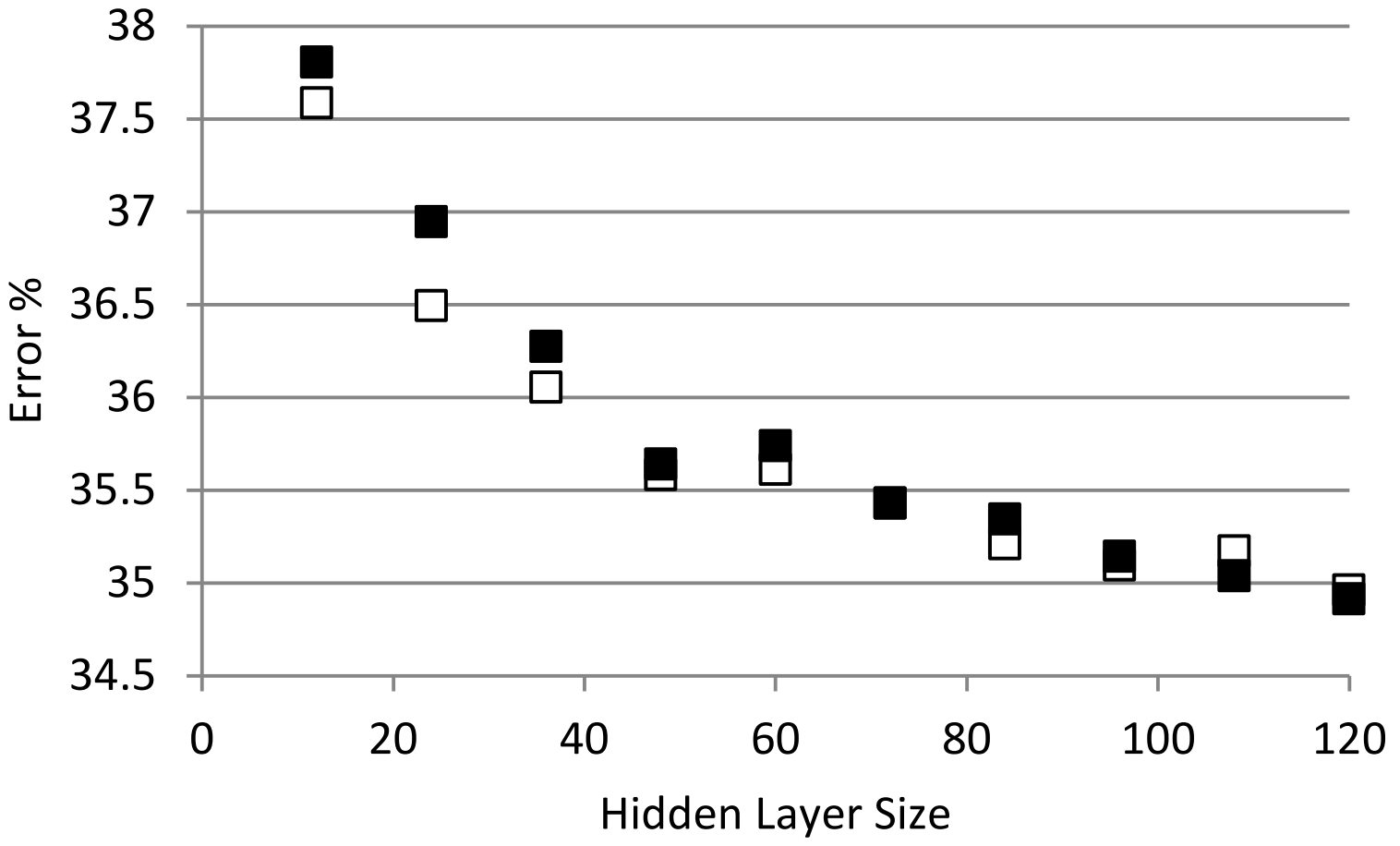

**Fig. 5.** Comparison of ELM (white squares) and CIW-ELM (black squares) performance on the *abalone* classification benchmark. It can be seen that there is little difference in performance (note the fineness of the vertical scale). This is well–known to be an underdetermined problem with the best results for all standard classification methods being no lower than 34%.

## 6 Conclusions

We have presented a schema for generation of input layer weights in ELM which offers reduced computation time and increased accuracy on some standard classification problems. The method is quick to implement and can be used in a deterministic and repeatable fashion. It still remains to evaluate this method on a wide range of benchmark problems and establish the circumstances under which it is a good choice for machine learning.